\documentclass[11pt,a4paper]{article}
\usepackage[hyperref]{eacl2021}
\usepackage{times}
\usepackage{latexsym}
\usepackage[textsize=tiny]{todonotes}
\usepackage{booktabs}
\usepackage{amsmath}
\usepackage{url}
\usepackage{xcolor}
\usepackage{fancyvrb}

\usepackage{float}

\usepackage{microtype}

\aclfinalcopy 
\def\aclpaperid{1394} 

\title{Metrical Tagging in the Wild: \\ Building and Annotating Poetry Corpora with Rhythmic Features}

\author{Thomas Haider \\
  Department of Language and Literature, \\
  Max Planck Institute for Empirical Aesthetics, Frankfurt am Main \\
  Institute for Natural Language Processing (IMS), \\
  University of Stuttgart \\
  \texttt{thomas.haider@ae.mpg.de}}

\date{}

\begin{document}
\maketitle
\begin{abstract}
A prerequisite for the computational study of literature is the availability of properly digitized texts, ideally with reliable meta-data and ground-truth annotation. 
Poetry corpora do exist for a number of languages, but larger collections lack consistency and are encoded in various standards, while annotated corpora are typically constrained to a particular genre and/or were designed for the analysis of certain linguistic features (like rhyme). 
In this work, we provide large poetry corpora for English and German, and annotate prosodic features in smaller corpora to train corpus driven neural models that enable robust large scale analysis. 

We show that BiLSTM-CRF models with syllable embeddings outperform a CRF baseline and different BERT-based approaches. In a multi-task setup, particular beneficial task relations illustrate the inter-dependence of poetic features. A model learns foot boundaries better when jointly predicting syllable stress, aesthetic emotions and verse measures benefit from each other, and we find that caesuras are quite dependent on syntax and also integral to shaping the overall measure of the line.

\end{abstract}

\section{Introduction}
\label{sec:intro}

Metrical verse, lyric as well as epic, was already common in preliterate cultures \cite{beissinger2012oral}, and to this day the majority of poetry across the world is drafted in verse \cite{fabb2008meter}.
In order to reconstruct such oral traditions, literary scholars mainly study textual
resources (rather than audio).
The rhythmical analysis of poetic verse is still widely carried out by example- and theory-driven manual annotation of experts, 
through so-called close reading \cite{carper2020meter,kiparsky2020metered,attridge2014rhythms,menninghaus2017emotional}.
Fortunately, well-defined constraints and the regularity of metrically bound language
aid the prosodic interpretation of poetry.

However, for projects that work with larger text corpora, close reading and extensive manual annotation are neither practical nor affordable. 
While the speech processing community explores end-to-end methods to detect and control the overall personal 
and emotional aspects of speech, including fine-grained features like pitch, tone, speech rate, cadence, and accent \cite{valle2020flowtron},
applied linguists and digital humanists still rely on rule-based tools \cite{plechavc2019relative,anttilagenre2016phonological,kraxenberger2016mimological}, some with limited generality \cite{navarro2018metrical}, 
or without proper evaluation \cite{bobenhausen2011metricalizer2}.
Other approaches to computational prosody are based on words in prose rather than syllables in poetry \cite{talman2019predicting,nenkova2007memorize}, rely on lexical resources with stress annotation such as the CMU dictionary, \cite{hopkins2017automatically,ghazvininejad2016generating}, are in need of an aligned audio signal \cite{rosenberg2010autobi,rosiger2015using}, or model only narrow domains such as iambic pentameter \cite{greene2010automatic,hopkins2017automatically,lau2018deep} or Middle High German \cite{estes2016supervised}.

To overcome these limitations, we propose corpus driven neural models to predict the prosodic features of syllables. These models should be evaluated against rhythmically diverse data, both on the level of the syllable and the whole line, i.e., whether the overall measure of a line is modeled correctly. 
Also, even though practically every culture has a rich heritage of poetic writing, large comprehensive collections are rare. 
We present datasets in German and English, encompassing a varied sample of around 7000 manually annotated lines, and we automatically tag large corpora in both languages to advance computational work on literature and rhythm. This may include the analysis and generation of poetry, but also more general work on prosody, or even speech synthesis.

Our main contributions are:
\begin{enumerate}
    \item The collection and standardization of heterogenous text sources that span writing of the last 400 years for both English and German, together comprising over 5 million lines of poetry.
    \item The annotation of prosodic features in a diverse sample of smaller corpora, including metrical and rhythmical features and the development of regular expressions to determine verse measure labels.
    \item The development of preprocessing tools and sequence tagging models 
    to jointly learn our annotations in a multi-task setup, highlighting the relationships of poetic features with each other. 
\end{enumerate}

\section{Manual Annotation}
\label{sec:resources}

We annotate prosodic features in two small poetry corpora that were previously collected and annotated for aesthetic emotions by \citet{haider2020po}. Both corpora cover a time period from around 1600 to 1930 CE, thus encompassing public domain literature from the modern period. The English corpus contains 64 poems with 1212 lines. The German corpus, after removing poems that do not permit a metrical analysis, 
contains 153 poems with 3489 lines in total. 
Both corpora are annotated with some metadata such as the title of a poem and the name and dates of birth and death of its author. The German corpus further contains annotation on the year of publication and literary periods. 

Figure \ref{fig:poetry} illustrates our annotation layers with three fairly common ways in which poetic lines can be arranged in modern English. A poetic line is also typically called \textit{verse}, from Lat.\ \textit{versus}, originally meaning to turn a plow at the ends of successive furrows, which, by analogy, suggests lines of writing \cite{steele2012verseandprose}.


In English or German, the rhythm of a linguistic utterance is basically determined by the sequence of syllable-related accent values (associated with pitch, duration and volume/loudness values) resulting from the `natural' pronunciation of a line, sentence, or text, by a competent speaker who takes into account the learned inherent word accents as well as syntax- and discourse-driven accents. Thus, lexical material comes with n-ary degrees of stress, depending on morphological, syntactic, and information structural context. The prominence (or stress) of a syllable is thereby dependent on other syllables in its vicinity, such that a syllable is pronounced relatively louder, higher pitched, or longer than its adjacent syllable. 

In this work, we manually annotate the sequence of syllables for metrical (meter, met) prominence (\texttt{+/-}), including a grouping of recurring metrical patterns, i.e., foot boundaries (\texttt{|}). We also operationalize a more natural speech rhythm (rhy) by annotating pauses in speech, caesuras (\texttt{:}), that segment the verse into rhythmic groups. In these groups we assign primary accents (2), side accents (1) and null accents (0). Also, we develop a set of regular expressions that derive the verse measure  (msr) of a line from its raw metrical annotation. 

\begin{figure}[htbp!]
\begin{footnotesize}
\begin{Verbatim}[commandchars=\\\{\},codes={\catcode`$=3\catcode`_=8}]
msr iambic.pentameter
met -  + | -   + | -   + |  -  +| -  +  | 
rhy 0  1   0   0   0   2 :  0  1  0  2  :
   My \textbf{love} is \textbf{like} to \textbf{ice}, and \textbf{I} to \textbf{fire}:

msr iambic.tetrameter
met  -   + |-   + | -   + | -   +  | 
rhy  0   1  0   2   0   0   0   1  :
    The \textbf{win}ter \textbf{eve}ning \textbf{set}tles \textbf{down}

msr trochaic.tetrameter
met  +    - | +   - | +  -  | + 
rhy  0    0   1   0   1  0    2  : 
    \textbf{Walk} the \textbf{deck} my \textbf{Cap}tain \textbf{lies},
\end{Verbatim}
\end{footnotesize}
\caption{Examples of rhythmically annotated poetic lines, with meter (\texttt{+/-}), feet (\texttt{|}), main accents (\texttt{2,1,0)}, caesuras (\texttt{:}), and verse measures (msr). Authors: Edmund Spenser, T.S. Eliot, and Walt Whitman.}
\label{fig:poetry}
\end{figure}

\subsection{Annotation Workflow}

Prosodic annotation allows for a certain amount of freedom of interpretation and (contextual) ambiguity, where several interpretations can be equally plausible. 
The eventual quality of annotated data can rest on a multitude of factors, such as the extent of training of annotators, the annotation environment, the choice of categories to annotate, and the personal preference of subjects \cite{mo2008naive,kakouros2016analyzing}.

Three university students of linguistics/literature were involved in the manual annotation process. They annotated 
by silent reading of the poetry, largely following an intuitive notion of speech rhythm, as was the mode of operation in related work \cite{estes2016supervised}.
The annotators additionally incorporated philological knowledge to recognize instances of poetic license, i.e., knowing how the piece is supposed to be read. 
Especially the annotation accuracy of metrical syllable stress and foot boundaries benefited from recognizing the schematic consistency of repeated verse measures, license through rhyme, or particular stanza forms.

\subsection{Annotation Layers}

In this paper, we incorporate both a linguistic-systematic and a historically-intentional analysis \cite{mellmann2versanalyse}, aiming at a systematic linguistic description of the prosodic features of poetic texts, but also using labels that are borrowed from historically grown traditions to describe certain forms or patterns (such as verse measure labels). 

We evaluated our annotation by calculating Cohen's Kappa between annotators. To capture different granularities of correctness, we calculated agreement on syllable level (accent/stress), between syllables (for foot or caesura), and on full lines (whether the entire line sequence is correct given a certain feature).

\paragraph{Main Accents \& Caesuras:}
Caesuras are pauses in speech. While a caesura at the end of a line is the norm (to pause at the line break) there are often natural pauses in the middle of a line. In few cases the line might also run on without a pause. As can be seen in Figure \ref{fig:poetry}, punctuation is a good signal for caesuras.  Caesuras (csr) are denoted with a colon.
We operationalize rhythm by annotating three degrees of syllable stress, so-called 'main accents' (m.ac), where the verse is first segmented into rhythmic groups by annotating caesuras, and in these groups we assign primary accents (2), side accents (1) and null accents (0).



 \begin{table}[ht]
      \centering
       \begin{tabular}{l|c|c|c|c|}
\toprule
   & \multicolumn{2}{c|}{Syllable} & \multicolumn{2}{c|}{Whole Line} \\
        &   m.ac   &   caesura  & m.ac   & caesura  \\
\midrule
$\mathrm{DE}_{blind}$   & .84           & .92            & .59           & .89       \\
\midrule
$\mathrm{EN}_{blind}$   & .80           & .88            & .66           & .86        \\
\bottomrule
\end{tabular}
  \caption{Cohen Kappa Agreement for Main Accents and Caesura}
  \label{agreement:rhythm}
    \end{table}
  
\begin{figure}[ht]
\centering
  \includegraphics[width=.8\linewidth]{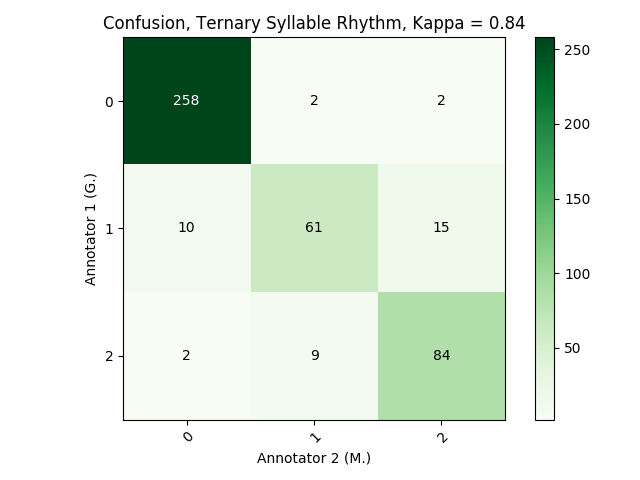}
  \caption{Confusion of German Main Accents}
  \label{fig:sub1}
\end{figure}

Six German and ten English poems were annotated by two annotators to calculate the agreement for rhythm. 
Table \ref{agreement:rhythm} lists the agreement figures for main accents (m.ac) and caesuras. It shows that caesuras can be fairly reliably detected through silent reading in both languages. On the other hand, agreement on main accents is challenging. Figure \ref{fig:sub1} shows the confusion of main accents for German. While 0s are quite unambiguous, it is not always clear when to set a primary (2) or side accent (1).

\paragraph{Meter and Foot:} 

In poetry, meter 
is the basic prosodic structure of a verse. 
The underlying abstract, and often top-down prescribed, meter consists of a sequence of beat-bearing units (syllables) that are either prominent or non-prominent. 
Non-prominent beats are attached to prominent ones to build metrical feet (e.g. iambic or trochaic ones). This metrical structure is the scaffold, as it were, for the linguistic rhythm.
Annotators first annotated the stress of syllables and in a subsequent step determined groupings of these syllables with foot boundaries, thus a foot is the grouping of metrical syllables. The meter (or measure) of a verse can be described as a regular sequence of feet, according to a specific sequence of syllable stress values, as elaborated further down on 'Verse Measures'.  



\begin{table}[htbp!]
      \centering
       \begin{tabular}{l|c|c|c|c|}
\toprule
   & \multicolumn{2}{c|}{Syllable} & \multicolumn{2}{c|}{Whole Line} \\
        &   meter   &   foot  & meter   & foot  \\
\midrule
$\mathrm{DE}_{corr.}$  & .98           & .87            & .94           & .71        \\
$\mathrm{DE}_{blind}$   & .98           & .79            & .92           & .71       \\
\midrule
$\mathrm{EN}_{blind}$   & .94           & .95            & .87           & .88        \\
\bottomrule
\end{tabular}
  \caption{Cohen Kappa Agreement for Metrical Stress and Foot Boundaries. Corr. is the agreement of the first version against the corrected version. Blind means that annotators did not see another annotation.}
  \label{agreement:total}
    \end{table}



The meter annotation for the German data was first done in a full pass by a graduate student. A second student then started correcting this annotation with frequent discussions with the first author. 
While on average the agreement scores for all levels of annotation suggested reliable annotation after an initial batch of of 20 German poems, we found that agreement on particular poems was far lower than the average, especially for foot boundaries. Therefore we corrected the whole set of 153 German poems, and the first author did a final pass. The agreement of this corrected version against the first version is shown in Table \ref{agreement:total} in the row $\mathrm{DE}_{corr.}$. To check whether annotators also agree when not exposed to pre-annotated data, a third annotator and the second annotator each annotated 10 diverse German poems from scratch. This is shown in $\mathrm{DE}_{blind}$.
For English, annotators 2 and 3 annotated 6 poems blind and then split the corpus.

Notably, agreement on syllables is acceptable, but feet were a bit problematic, especially for German. To investigate the sources of disagreement, we calculated agreement of feet on all 153 poems. Close reading for disagreement of foot boundaries revealed that poems with $\kappa$ around .8 had faulty guideline application (annotation error). 14 poems had an overall $\kappa$ $<$ .6, which stemmed from either ambiguous rhythmical structure (multiple annotations are acceptable) and/or schema invariance, where a philological eye considers the whole structure of the poem and a naive annotation approach does not render the intended prosody correctly.   

As an example for ambiguous foot boundaries, the following poem, Schiller's `Bürgschaft', can be set in either \textit{amphibrachic} feet, or as a mixture of \textit{iambic} and \textit{anapaestic} feet. Such conflicting annotations were discussed by \newcite{heyse1827theoretisch}, who finds that in the Greek tradition the \textit{anapaest} is preferable, but a `weak amphibrachic gait' allows for a freer rhythmic composition. Thus, we suggest that Schiller was breaking with tradition. 

\begin{scriptsize}
\begin{Verbatim}[commandchars=\\\{\},codes={\catcode`$=3\catcode`_=8}]
(Foot Boundary Ambiguity) Schiller, \textit{'Die Bürgschaft'}

(1) met="-+-|-+-|-+-|"     
    Ich \textbf{las}se |    den \textbf{Freund}   dir | als \textbf{Bür}gen, |
(2) met="-+|--+|--+|-"     
    Ich \textbf{las}   | se den \textbf{Freund} | dir   als \textbf{Bürg} | en,
Transl.: I leave this friend to you as guarantor
        
(1) met="-+-|-+-|-+-|"     
    Ihn \textbf{magst}   du, | ent\textbf{rinn'}   ich, | er\textbf{wür}gen. |
(2) met="-+|--+|--+|-"     
    Ihn \textbf{magst} | du,   ent\textbf{rinn'} | ich,   er\textbf{wür} | gen.
Transl.: Him you may strangle if I escape.
        
(1) (amphibrach)
(2) (iambus / anapaest)
\end{Verbatim}
\end{scriptsize}

\paragraph{Verse Measures:} 

We develop a set of regular expressions to determine the measure of a line from its raw metrical annotation.  We orient ourselves with the handbook of \newcite{knorrich1971deutsche}. The `verse measure' (msr) is a label for the whole line according to recurring metrical feet. We label the verse according to its dominant foot, i.e., the repetition of patterns like \texttt{iambus} (\texttt{-+}), \texttt{trochee} (\texttt{+-}), \texttt{dactyl} (\texttt{+--}), \texttt{anapaest} (\texttt{--+}), or \texttt{amphibrach} (\texttt{-+-}). Also, the rules determine the number of stressed syllables in the line, where \textit{di-, tri-, tetra-, penta-}, and \textit{hexameter} signify 2, 3, 4, 5, and 6 stressed syllables accordingly. Thus, \texttt{+-+-+-} is an example for a trochaic.trimeter and \texttt{-+-+-+-+} is a iambic.tetrameter, since the foot boundaries should look like this: \texttt{-+|-+|-+|-+|}. Typically, female (unstressed) line endings are optional (cadence).
Additionally, we annotate labels for (i) \texttt{inversion}, when the first foot is inverted, e.g., the first foot in a iambic line is trochaic: \texttt{+--+-+-+}, (ii) \texttt{relaxed}, if an unstressed syllable was inserted: \texttt{-+-+--+-+} (iambic.tetrameter.relaxed), (iii) and choliambic endings: \texttt{-+-+-+--+}. 
Besides these basic forms, we also implement historically important forms such as a strict \texttt{alexandrine},\footnote{Alexandrine: \texttt{-+-+-+-+-+-+-?} \\ The symbol before ? is optional} 
the \texttt{dactylic hexameter},\footnote{Hexameter: \texttt{+--?+--?+--?+--?+--+-}} conventionally known as `hexa\-meter', and some ode forms like the \texttt{asklepiadic} verse (\texttt{+-+--++--+-+}).


Table \ref{tab:measures} lists the most frequent labels for each language without length, called short measure (smsr). The English data includes all datasets that are used in the experiments, as discussed in section \ref{sec:dataformat}.

\begin{table}[htbp!]
    \centering
    \begin{footnotesize}
    \begin{tabular}{r|r||r|r|}
    \multicolumn{2}{c||}{English} & \multicolumn{2}{c|}{German} \\
    \midrule
    smsr   & freq. & smsr & freq.\\
    \midrule
     iambic     & 2096      & iambic & 1976 \\
     trochaic   & 490       & trochaic & 793 \\
     anapaest   & 306       & amphibrach & 258 \\
     amphibrach & 255       & alexandrine & 206 \\
     daktylic   & 248       & daktylic & 76 \\
     hexameter  & 152       & anapaest & 72 \\
     prosodiakos & 91       & asklepiade & 26 \\
     other      & 52        & pherekrateus & 17 \\
     alexandrine& 35        &  glykoneus & 14 \\
    \end{tabular}
    \caption{Most frequent verse measures in small English and German corpora, without length.}
    \label{tab:measures}
    \end{footnotesize}
\end{table}

\section{Large Poetry Corpora}
\label{sec:largecorp}

In order to enable large scale experiments on poetry, we collect and standardize large poetry corpora for English and German. The English corpus contains around 3 million lines, while the German corpus contains around 2 million lines. 
The corpora and code can be found at \url{https://github.com/tnhaider/metrical-tagging-in-the-wild}



Our resources are designed in a standardized format to sustainably and interoperably archive poetry in both .json and TEI P5 XML. 
The .json format is intended for ease of use and speed of processing while retaining some expressiveness. Our XML format is built on top of a ``Base Format", the so-called DTA-Basisformat\footnote{\url{http://www.deutschestextarchiv.de/doku/basisformat/}} \cite{haaf2014dta} that not only constrains the data to TEI P5 guidelines, 
but also regarding a stricter relaxNG schema that we modified for our annotation.\footnote{This schema defines a strict layout of poetic annotation. It allows us to validate XML files regarding their correctness. It is thus useful for manual annotation with the OxygenXML editor, avoiding parsing errors downstream.} 

\subsection{A Large German Poetry Corpus}

\begin{figure}[htbp]
    \hspace*{-0.5cm}\includegraphics[width=1.05\linewidth]{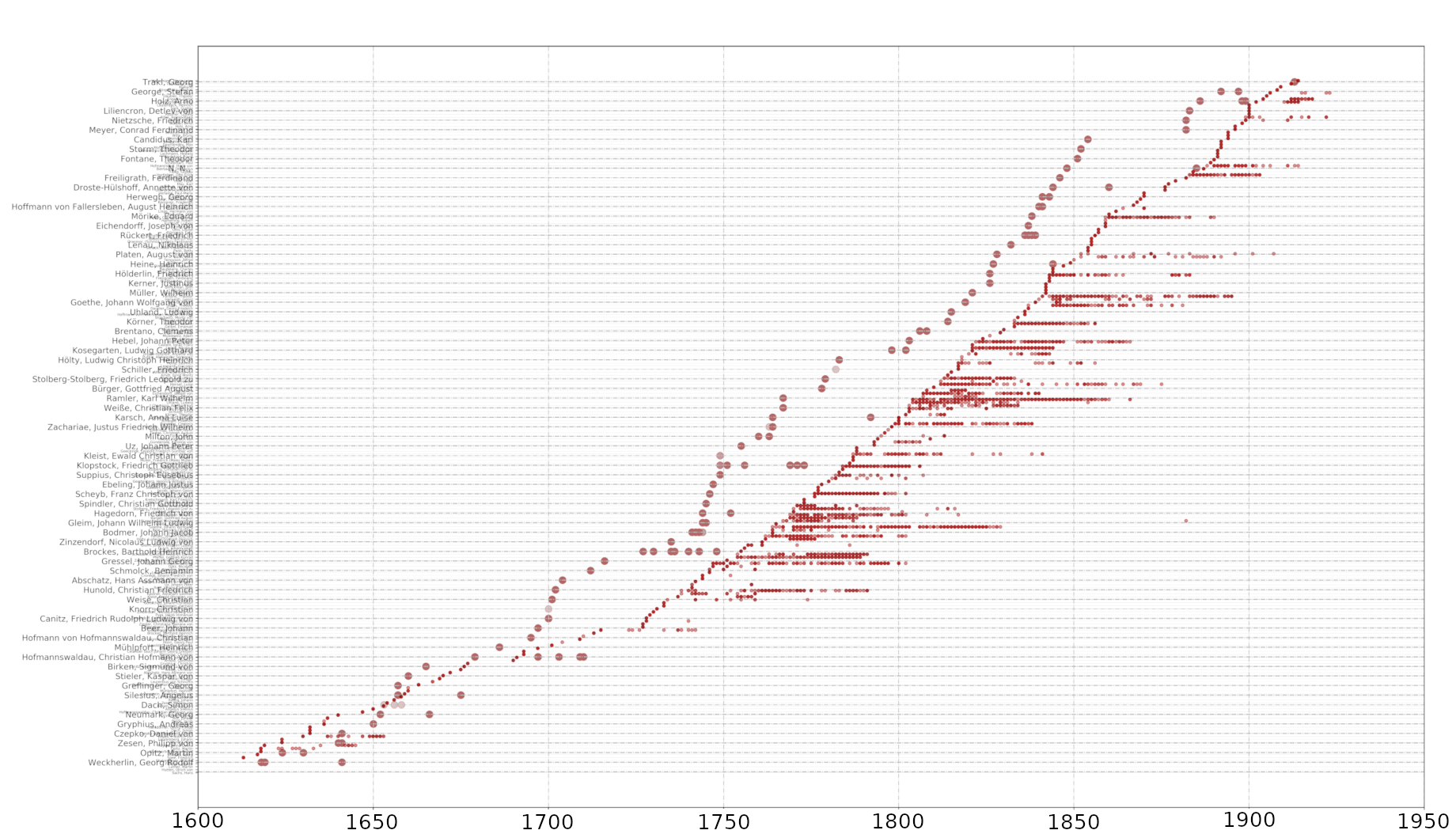}
    \caption{Each dot represents a poem of a German Author (y-axis) over Time (x-axis) from Textgrid (small dots bottom) and DTA (large dots top). 1600--1950. Authors are not aligned, and poems can be on top of each other.}
    \label{fig:authorlife}
\end{figure}

We built a large, comprehensive, and easily searchable resource of New High German poetry by collecting and parsing the bulk of digitized corpora that contain public domain German literature. This includes the German Text Archive (DTA) (\url{http://deutschestextarchiv.de}) the Digital Library of Textgrid (\url{http://textgrid.de}), and also the German version of Project Gutenberg (which we omit from our experiments due to inconsistency).\footnote{\url{https://www.projekt-gutenberg.org/}}

Each of these text collections is encoded with different conventions and varying degrees of consistency. Textgrid contains 51,264 poems with the genre label `Verse', while DTA contains 23,877 poems with the genre label `Lyrik'. It should be noted that the whole DTA corpus contains in total 40,077 line groups that look like poems, but without the proper genre label, poems are likely embedded within other texts and might not come with proper meta-data. We implement XML parsers in python 
to extract each poem with its metadata and fix stanza and line boundaries. The metadata includes the author name, the title of the text, the year it was published, the title and genre of the volume it was published in, and finally, an identifier to retrieve the original source.
We perform a cleaning procedure that removes extant XML information, obvious OCR mistakes, and normalize umlauts and special characters in various encodings,\footnote{
We fix the orthography both on string and bytecode level. We replace the rotunda (U+A75B) and the long s (U+017F), the latter of which is pervasive in DTA.} particularly in DTA.
We use langdetect\footnote{\url{https://pypi.org/project/langdetect/}} 1.0.8 to tag every poem with its language to filter out any poems that are not German (such as Latin or French). 
The corpus finally contains almost 2M lines in over 60k poems. In Figure \ref{fig:authorlife} we plotted each poem in DTA and Textgrid over time, from 1600 to 1950. The x-axis shows the year of a poem, while the y-axis is populated by authors. One can see that DTA consists of full books that are organized by author (large dots) so that the datapoints for single poems get plotted on top of each other, while Textgrid has a time stamp for most single poems (after 1750), outlining the productive periods of authors. 

\subsection{A Large English Poetry Corpus}

The English corpus contains the entirety of poetry that is available in the English Project Gutenberg (EPG) collection. We firstly collected all files with the metadatum `poetry' in (temporal) batches with the GutenTag tool \cite{brooke2015gutentag}, to then parse the entire collection in order to standardize the inconsistent XML annotation of GutenTag and remove duplicates, since EPG contains numerous different editions and issues containing the same material. We also filter out any lines (or tokens) that indicate illustrations, stage directions and the like. We use langdetect 
to filter any non-English material.

The github repository of \citet{aparrish} previously provided the poetry in EPG by filtering single lines with a simple heuristic (anything that could look like a line), not only including prose with line breaks, but also without conserving the integrity of poems but providing a document identifier per line to find its origin. We offer our corpus in XML with intact document segmentation and metadata, still containing over 2.8 million lines.

\section{Experiments}

In the following, we carry out experiments to learn/model the previously annotated features and determine their degree of informativeness for each other with a multi-task setup. We include two additional datasets with English meter annotation, and evaluate pre-processing models for syllabification and part-of-speech tagging.

\subsection{Preprocessing} 

Tokenization for both languages is performed with SoMaJo \cite{Proisl_Uhrig_EmpiriST:2016}, with a more conservative handling of apostrophes (to leave words with elided vowels intact). This tokenizer is more robust regarding special characters than NLTK. 
We also train models for hyphenation (syllabification) and part-of-speech (POS) tagging, since syllabification is a prerequisite to analyse prosody, and POS annotation allows us to gauge the role of syntax for prosodic analysis.

\subsubsection{Hyphenation / Syllabification}

For our purposes, proper syllable boundaries are paramount to determine the segmentation of lines regarding their rhythmic units. 
We test the following systems:\footnote{Syllabipy determines boundaries based on the sonority principle, Pyphen uses the Hunspell dictionaries, and HypheNN is a simple feed forward network that is trained on character windows.} \textit{Sonoripy},\footnote{\url{https://github.com/alexestes/SonoriPy} \url{https://github.com/henchc/syllabipy}} \textit{Pyphen},\footnote{\url{pyphen.org}} \textit{hypheNN},\footnote{\url{github.com/msiemens/HypheNN-de}} and a BiLSTM-CRF \cite{Reimers:2017:EMNLP}\footnote{\url{https://github.com/UKPLab/emnlp2017-bilstm-cnn-crf}} with pretrained word2vec character embeddings. These embeddings were trained on the corpora in section \ref{sec:largecorp}.

To train and test our models, we use CELEX2 for English and extract hyphenation annotation from wiktionary for German.\footnote{For German, wiktionary contains 398.482 hyphenated words, and 130.000 word forms in CELEX. Unfortunately, German CELEX does not have proper umlauts, and models trained on these were not suitable for poetry. For English, wiktionary only contains 5,142 hyphenated words, but 160,000 word forms in CELEX.} We evaluate our models on 20,000 
randomly held-out words for each language on word accuracy and syllable count. Word accuracy rejects any word with imperfect character boundaries, while syllable count is the more important figure to determine the proper length of a line. As seen in Table \ref{tab:syllabel_eval}, the BiLSTM-CRF performs best for English and does not need any postprocessing. For German, the LSTM model is less useful as it tends to overfit, where over 10\% of annotated lines were still rejected even though in-domain evaluation suggests good performance. We therefore use an ensemble with HypheNN, Pyphen and heuristic corrections for German, with only 3\% error on the gold data, as seen in Table \ref{table:gold_syllables} (the datasets are discussed in section \ref{sec:dataformat}). 

\begin{table}[htbp!]
\centering
\begin{small}
\begin{tabular}{r|c|c|c|c|}
& \multicolumn{2}{c|}{German} & \multicolumn{2}{c|}{English} \\
& w. acc. & sy. cnt  & w. acc.  & sy. cnt  \\
\midrule
SonoriPy &  .476 & .872 & .270 & .642 \\
Pyphen &  .839  & .875 & .475 & .591 \\
HypheNN & .909 & .910 & .822 & .871 \\
BiLSTM-CRF & \textbf{.939} & \textbf{.978} & \textbf{.936} & \textbf{.984} \\
\end{tabular}
\caption{Evaluation of Syllabification Systems on Wiktionary (German) and CELEX (English).}
\label{tab:syllabel_eval}
\end{small}
\end{table}

\begin{table}[htbp!]
    \centering
    \begin{footnotesize}
    \begin{tabular}{lrrrr} 
        \toprule 
         & German & EPG64 & FORB & PROS \\ 
         \cmidrule(r){1-1}\cmidrule(rl){2-2}\cmidrule(l){3-3}\cmidrule(l){4-4}\cmidrule(l){5-5}
         \# correct lines & 3431 & 1098   & 1084 & 1564 \\
         \# faulty lines & 58 & 114      & 49 &  173 \\
         \bottomrule
    \end{tabular}
    \caption{Size of manually annotated corpora. Faulty lines denotes the number of lines where the automatic syllabification failed. Correct lines are used for experiments, since only there the gold annotation aligns.}
    \label{table:gold_syllables}
    \end{footnotesize}
\end{table}

\begin{table*}[htbp]
    \centering
    \begin{small}
    \begin{tabular}{l|cccccccccc}
    \toprule
    POS           & Noun & Adj. & Full V. & Adverb & Modal V. & Interj. & Pron. & Prep. & Konj. & Art. \\
    \midrule
    Abr. POS Tag  &  NN  &  ADJ & VV & ADV  & VM & ITJ    & P   & AP  & KO    & AR \\
    \midrule
    Accent Ratio  & .97  &  .89 & .84 & .75 & .73 & .55  & .4--.015 & .27 & .23 & .06 \\
    \bottomrule
    \end{tabular}
    \end{small}
    \caption{Accent ratio for part-of-speech of German monosyllabic words (ratio of metrical stress).}
    \label{tab:posstress}
\end{table*}

\subsubsection{POS tagging}\label{sec:postagging}

Since we are dealing with historical data, POS taggers trained on current data might degrade in quality and it has been frequently noted that poetry makes use of non-canonical syntactic structures \cite{gopidi2019computational}. For German, we evaluate the robustness of POS taggers across different text genres. We use the gold annotation of the TIGER corpus (modern newspaper), and pre-tagged sentences from DTA, including annotated poetry (Lyrik), fiction (Belletristik) and news (Zeitung).\footnote{DTA was tagged with TreeTagger and manually corrected afterwards. \url{http://www.deutschestextarchiv.de/doku/pos}} The STTS tagset is used. We train and test Conditional Random Fields (CRF)\footnote{From the sklearn crf-suite} to determine a robust POS model.\footnote{As features, we use the word form, the preceding and following two words and POS tags, orthographic information (capitalization), character prefixes and suffixes of length 1, 2, 3 and 4.} See Table \ref{tab:poseval} for an overview of the cross-genre evaluation. We find that training on TIGER is not robust to tag across domains, falling to around .8 F1-score when tested against poetry and news from DTA. These results suggest that the loss in tagging accuracy is mainly due to (historical) orthography, and to a lesser extent due to local syntactic inversions. 

 \begin{table}[htbp!]
 \center 
 \begin{scriptsize}
\begin{tabular}{l||c|c|c|c|c}
Test & \multicolumn{4}{c}{Train}  \\
\toprule
            & TIGER         & DTA   & DTA+TIG. & Belletr. & Lyrik  \\
\toprule
Lyrik       & .795 & .949  & .948      & .947          & \textbf{.953} \\
Belletristik& .837          & \textbf{.956}  & .954      & .955          & .955 \\
DTA Zeitung & .793          & \textbf{.934}  & .933      & .911          & .900 \\
TIGER       & \textbf{.971}          & .928  & .958      & .929          & .913 \\ 
\bottomrule
\end{tabular}
\caption{Evaluation of German POS taggers across genres. F1-scores.}
    \label{tab:poseval}
\end{scriptsize}
\end{table}

For English, we test the Stanford core-nlp tagger.\footnote{\url{https://nlp.stanford.edu/software/tagger.shtml}} The tagset follows the convention of the Penn TreeBank. This tagger is not geared towards historical poetry and consequently fails in a number of cases. We manually correct 50 random lines and determine an accuracy of 72\%, 
where particularly the `NN' tag is overused. This renders the English POS annotation unreliable for further experiments. 


\subsubsection{Additional Data and Format}
\label{sec:dataformat}

The annotated corpora for English include: (1) The for-better-for-verse collection (FORB)\footnote{\url{https://github.com/manexagirrezabal/for_better_for_verse/tree/master/poems}} with around 1200 lines, which was used by \newcite{agirrezabal-etal-2016-machine,agirrezabal2019comparison}, and (2) the 1700 lines of poetry against which \texttt{prosodic}\footnote{\url{https://github.com/quadrismegistus/prosodic}} \cite{algee2014stanford} was evaluated (PROS). We merge these with (3) our own 1200 lines in 64 English poems (EPG64). The first two corpora were already annotated for metrical syllable stress.  However, FORB does not contain readily available foot boundaries, and in PROS, foot boundaries are occasionally set after each syllable.\footnote{Additionally, FORB makes use of a \texttt{<seg>} tag to indicate syllable boundaries, so we do not derive the position of a syllable in a word. It also contains two competing annotations, \texttt{<met>} and \texttt{<real>}. The former is the supposedly proper metrical annotation, while the latter corresponds to a more natural rhythm (with a tendency to accept inversions and stress clashes). We only chose \texttt{<real>} when \texttt{<met>} doesn't match the syllable count (ca. 200 cases), likely deviating from the setup in \cite{agirrezabal-etal-2016-machine,agirrezabal2019comparison}.}
Table \ref{table:gold_syllables} shows the number of lines that were correctly syllabified (number of syllables of metrical annotation) by our best systems. Correctly segmented lines are used in the experiments, but we discard faulty lines to avoid improper feature alignment.


Figure \ref{fig:exanno} shows the data format of an example line. `Measure' is derived from the meter line via regular expressions. `Syll' is the position of the syllable in a word, 0 for monosyllaba, otherwise index starting at 1. We removed punctuation to properly render line measures, even though punctuation is a good signal for caesuras (see Figure \ref{fig:poetry}).

\begin{figure}[htbp!]
\begin{tiny}
\begin{Verbatim}[commandchars=\\\{\},codes={\catcode`$=3\catcode`_=8}]
# tok  met ft pos syll csr main smsr    measure    met\_line

1  Look  +  .  VB   0   .  1  iambic  i.penta.inv +--+-+-+-+
2  on    -  .  IN   0   .  0  iambic  i.penta.inv +--+-+-+-+
3  my    -  .  PRP\$ 0   .  0  iambic  i.penta.inv +--+-+-+-+
4  works +  :  NNS  0   :  2  iambic  i.penta.inv +--+-+-+-+
5  ye    -  .  PRP\$ 0   .  0  iambic  i.penta.inv +--+-+-+-+
6  Might +  :  NNP  1   .  1  iambic  i.penta.inv +--+-+-+-+
7  y     -  .  NNP  2   :  0  iambic  i.penta.inv +--+-+-+-+
8  and   +  :  CC   0   .  0  iambic  i.penta.inv +--+-+-+-+
9  de    -  .  VB   1   .  0  iambic  i.penta.inv +--+-+-+-+
10 spair'+  :  VB   2   :  1  iambic  i.penta.inv +--+-+-+-+
\end{Verbatim}
\end{tiny}
\caption{Tabular data format for experiments. Author of this line: Percy Blythe Shelley.}
\label{fig:exanno}
\end{figure}

\subsection{Accent Ratio of Part-of-Speech}\label{sec:poshier}

Previous research has noted that part-of-speech annotation provides a good signal for the stress of words \cite{nenkova2007memorize,greene2010automatic}. To test this, we calculate the pos-accent ratio of monosyllabic words in our German annotation by dividing how often a particular part-of-speech appears stressed (\texttt{+}) in the corpus by how often this part-of-speech occurs in the corpus. We restrict this to monosyllabic words, as polysyllabic words typically have a lexical stress contour. The result is a hierarchy of stress that we report in Table \ref{tab:posstress}. At the ends of the spectrum, we see that nouns are usually stressed, while articles are seldom stressed.

\subsection{Learning Meter}

To learn the previously annotated metrical values for each syllable, the task is framed as sequence classification. Syllable tokens are at the input and the respective \texttt{met} labels at the output. We test a nominal CRF (see section \ref{sec:postagging}) and a BERT model as baselines and implement a BiLSTM-CRF\footnote{\url{https://github.com/UKPLab/emnlp2017-bilstm-cnn-crf}} with pre-trained syllable embeddings. These embeddings were trained by splitting all syllables in the corpora from section \ref{sec:largecorp}, and training word2vec embeddings over syllables. This system uses three layers of size 100 for the BiLSTM and does the final label prediction with a linear-Chain CRF. Variable dropout of .25 was applied at both input and output. No extra character encodings were used (as these hurt both speed and accuracy). 

We do a three fold cross validation with 80/10/10 splits and average the results, reporting results on the test set in Table \ref{tab:bestclass}. We evaluate prediction accuracy on syllables and the accuracy of whether the whole line was tagged correctly (line acc.). Line accuracy is especially important if we want to classify poetic verse measures.

 \begin{table}[htbp!]
    \centering
    \begin{scriptsize}
    \begin{tabular}{l||c|c|c|c|}
           & \multicolumn{2}{c|}{English}  & \multicolumn{2}{c|}{German} \\
                &  syll. acc    &  line acc     &  syll. acc    &  line acc     \\
                \midrule
    CRF         &  .922         &  .478         &  .941      &  .553 \\
    BERT        &  .850         &  .371         &  .932         &  .498 \\
    BiLSTM-CRF  &  \textbf{.955}         &  \textbf{.831}         &  \textbf{.968}         &  \textbf{.877} \\
    \midrule
    \midrule 
    Agirrezabal (2019) & .930   &  .614         &     -          & -\\
    Anttila \& Heuser (2016) & .894 & .607 & - & - \\
    \end{tabular}
    \caption{Best Classifiers for Metrical Syllable Stress}
    \label{tab:bestclass}
    \end{scriptsize}
\end{table}

Though not directly comparable (data composition differs), we include results as reported by \newcite{agirrezabal2019comparison} for the English for-better-for-verse dataset. We also test the system `\texttt{prosodic}' of \newcite{anttilagenre2016phonological} against our gold data (EPG64), resulting in .85 accuracy for syllables and .44 for lines. When only evaluating on lines that were syllabified to the correct length (their syllabifier), 27\% of lines are lost, but on this subset it achieves .89 syllable and .61 line accuracy.

 
Learning the sequence of metrical syllable stress with BERT cannot compete against our other models, possibly resulting from an improper syllable representation, as the word-piece tokenizer segments word chunks other than syllables. 

We also experiment with framing the task as document (line) classification, where BERT should learn the verse label (e.g., iambic.pentameter) for a given sequence of words. On the small English dataset, BERT only achieves around .22 F1-macro and .42 F1-micro (line F1). 
We then tagged 20,000 lines of the large English corpus with a BiLSTM-CRF model and trained BERT on this larger dataset, reaching .48 F1-macro and .62 F1-micro. In this setup, BERT detects frequent classes like iambic.pentameter or trochaic.tetrameter fairly well (.8), but it appears that this model mainly picks up on the length of lines and fails to learn measures other than iambus and trochee, like dactyl or anapaest, or irregular verse with inversions. This might limit experiments with transfer learning of verse measure knowledge.

\subsection{Pairwise Joint Prosodic Task Learning}

With the aim of learning the relationships between our different annotation layers, we performed experiments with a multi-task setup. We used the BiLSTM-CRF architecture from the previous experiment, where a sequence of syllable embedding vectors is at the input, and the respective sequence of labels at the output. We used the German dataset here, as the annotation is generally more reliable (e.g., POS). In this experiment we also try to learn the annotation of aesthetic emotions that was described for this dataset by \newcite{haider2020po}. Each line was annotated with one or two emotions from a set of nine emotions. Here, we only used the primary emotion label per line.

First, we trained a single task model for each annotation layer (\texttt{single}), then all tasks jointly (\texttt{+all}), and finally pair-wise combinations  (\texttt{+<auxiliary task>}). In Table \ref{tab:my_mtl}, we report the accuracy on syllable level for each main task, jointly learned with its respective auxiliary task. 


\begin{table}[htbp!]
    \centering
    \begin{scriptsize}
    \begin{tabular}{l||c|c|c|c|c|c|c|}
            & met  & feet& syllin  & pos    & csra  &  m.ac & emo \\
    \midrule
single    & .964& .871& \textbf{.952} &  .864   & .912   & .866 & .328 \\
    \midrule
+met & -   & \textbf{.922}& .949    & .856   &  .918  & .869 & .347   \\
    +feet & .961 & -   & .948   & .853    &  .917  & .863  & .369  \\
    +syllin  & .966 & .900& -      & .860   &  .919  & .867  & .330  \\
+pos & .956 & .879& \textbf{.953}   &  -     & \textbf{.924}   & \textbf{.879}   & \textbf{.393} \\
    +csra    & .961 & .886& .940   & .855   &  -     & .868 & .364  \\
    +m.ac  & .964 & \textbf{.915} &  .948   & .865    & .915   & - & .354   \\
    +smsr  & .965 & .884& .942   & .854     &  .918  & .868 & \textbf{.378}   \\
+fmsr& \textbf{.968}& .899 &  .938  &  .858      &  \textbf{.926}      & .868   & \textbf{.395} \\
+m\_line & .966    & .882   & .937       &  .853 &  .919      & .868 & \textbf{.398}  \\
+emo & .962   & .863  & .944       & .854   &  .921   & .864 &  - \\
    \midrule
    +all     & .967 & \textbf{.930} & .947  & .790  & .919   & .870 & .377 \\
    \end{tabular}
    \caption{Accuracy for Pairwise Joint Task Learning.}
    \label{tab:my_mtl}
    \end{scriptsize}
\end{table}

Note that learning syllable-level POS does not benefit from any other task, not even the syllable position in the word, while several tasks like caesuras, main accents and emotions benefit from additional POS information.  
Predicting meter also degrades from an additional POS task, which possibly interfers with the syllable embeddings. Meter might be also more contextual than suggested in Table \ref{tab:posstress}. 

However, meter tagging slightly benefits from fine-grained verse measure labels.
Interestingly, learning foot boundaries heavily benefits from jointly learning syllable stress. In a single task setup, foot boundaries are learned with .871 accuracy, but in combination with metrical stress, feet are learned with .922 acc. and in combination with main accents at .915. This might be expected, as foot groupings are dependent on the regularity of repeating metrical syllable stresses (though less dependent on main accents). However, our annotators only achieved Kappa agreement of .87 for feet. It is curious then, how the model overcomes this ambiguity. When learning all tasks jointly (\texttt{+all}), foot prediction even reaches .930, suggesting that feet are related to all other prosodic annotations.

We observe that the exchange between caesuras and main accents is negligible. However, caesuras benefit from POS (despite the absence of punctuation), syllable position (syllin) and global measures (msr), indicating that caesuras are integral to poetic rhythm and fairly dependent on syntax. 

For emotions we find, despite the hard task (line instead of stanza), and only using syllable embeddings rather than proper word embeddings, that the single task setup is already better than the majority baseline. More importantly, we can see that jointly learning POS or verse measure benefits the emotion prediction (slightly the meter prediction itself: .97). This suggests that there might be a systematic relationship between meter and emotion.

\section{Related Work}
\label{sec:related}

\subsection{Annotation of Prosodic Features}

Earlier work \cite{nenkova2007memorize} already found strong evidence that part-of-speech tags, accent-ratio\footnote{The ratio of how often a word form appears stressed vs. unstressed in a corpus} and local context provide good signals for the prediction of word stress. Subsequently, architectures like MLP \cite{agirrezabal-etal-2016-machine}, CRFs and LSTMs \cite{estes2016supervised,agirrezabal2019comparison} and transformer models \cite{talman2019predicting} have notably improved the performance to predict the prosodic stress of words and syllables. Unfortunately, most of this work only evaluates model accuracy on syllable or word level, with the exception of \newcite{agirrezabal2019comparison}.



A digital resource with annotation of poetic meter was missing for New High German. For Middle High German, \newcite{estes2016supervised} annotated a metrical scheme for hybrid meter. \newcite{anttila2018sentence} annotated main accents in political speeches. \newcite{agirrezabal-etal-2016-machine,agirrezabal2019comparison} used the English \textit{for-better-for-verse} and the dataset of \newcite{navarro2016metrical}, who annotated hendecasyllabic verse (11 syllables) in Spanish Golden Age sonnets. \newcite{algee2014stanford} annotated 1700 lines of English poetry to evaluate their system.

\subsection{Poetry Corpora \& Generation}

Several poetry corpora have been used in the NLP community. Work on English has strongly focused on iambic pentameter, e.g., of Shakespeare \cite{greene2010automatic} or with broader scope \cite{jhamtani2017shakespearizing,lau2018deep,hopkins2017automatically}. Other work has focused on specific genres like Spanish sonnets \cite{ruiz2020diachronic}, limericks \cite{jhamtani2019learning}, or Chinese Tang poetry \cite{zhang2014chinese}. There are further resources with rhyme patterns \cite{reddy2011unsupervised,haider2018supervised} or emotion annotation \cite{haider2020po}. 
Truly large corpora are still hard to find, besides the Gutenberg project for English and Textgrid and DTA for German. 

\section{Conclusion}

We created large poetry corpora for English and German to support computational literary studies, and annotated prosodic features in smaller corpora. Our evaluation shows that a multitude of features can be annotated through silent reading, including meter, main accents and caesuras, even though foot annotation can be challenging. Finally, we performed first experiments with a multi-task setup to find beneficial relations between certain prosodic tasks. Learning metrical annotation, including feet and caesuras, largely benefits from a global verse measure label, while foot boundaries also benefit from any joint learning with syllable stress and all features alltogether, even surpassing the human upper bound. In the future, this work will hopefully aid research on the link between rhythm and emotions that are elicited through poetry reading. 

\section*{Acknowledgments}

We thank Gesine Fuhrmann and Debby Trzeciak for their annotations. We thank Steffen Eger, Winfried Menninghaus, Anne-Kathrin Schumann, and our reviewers for feedback on the manuscript. Thanks also goes to Ryan Heuser for help with his tool \texttt{prosodic}.

\bibliography{eacl2021}
\bibliographystyle{acl_natbib}

\end{document}